\documentclass{ecai}
\usepackage{graphicx}
\usepackage{latexsym}
\usepackage{multirow}
\usepackage{multicol}
\usepackage{caption}
\usepackage{float}
\usepackage{tikz}
\usepackage{pythonhighlight}
\restylefloat{table}

\def\checkmark{\tikz\fill[scale=0.4](0,.35) -- (.25,0) -- (1,.7) -- (.25,.15) -- cycle;} 

\newcommand{\specialcell}[2][c]{ \begin{tabular}[#1]{@{}c@{}}#2\end{tabular}}

\begin{document}

\begin{frontmatter}

\title{Ensembling Uncertainty Measures \\ to Improve Safety of Black-Box Classifiers}

\author[A]{\fnms{Tommaso}~\snm{Zoppi}\orcid{0000-0001-9820-6047}\thanks{Corresponding Author. Email: tommaso.zoppi@unifi.it.}}
\author[A]{\fnms{Andrea}~\snm{Ceccarelli}\orcid{0000-0002-2291-2428}}
\author[A]{\fnms{Andrea}~\snm{Bondavalli}\orcid{0000-0001-7366-6530}}

\address[A]{Department of Mathematics and Informatics, University of Florence - Viale Morgagni 65, 50134 Florence, Italy}

\begin{abstract}
Machine Learning (ML) algorithms that perform
classification may predict the wrong class, experiencing
misclassifications. It is well-known that misclassifications may have
cascading effects on the encompassing system, possibly resulting in
critical failures. This paper proposes SPROUT, a Safety wraPper thROugh
ensembles of UncertainTy measures, which suspects misclassifications by
computing uncertainty measures on the inputs and outputs of a black-box
classifier. If a misclassification is detected, SPROUT blocks the
propagation of the output of the classifier to the encompassing system.
The resulting impact on safety is that SPROUT transforms erratic outputs
(misclassifications) into data omission failures, which can be easily
managed at the system level. SPROUT has a broad range of applications as it
fits binary and multi-class classification, comprising image and tabular
datasets. We experimentally show that SPROUT always identifies a huge
fraction of the misclassifications of supervised classifiers, and it is
able to detect all misclassifications in specific cases. SPROUT
implementation contains pre-trained wrappers, it is publicly available
and ready to be deployed with minimal effort.
\end{abstract}

\end{frontmatter}

\section{Introduction}

A typical approach to guarantee safety {[}40{]} is to equip a functional
component with a detector {[}30{]}, {[}33{]}, {[}36{]}, {[}37{]} so to
trigger a fail-safe or fail-stop behavior whenever the correct
functioning is not guaranteed. At the system level, it is often
desirable that safety-critical functions would either i) deliver a
correct result or ii) omit outputs i.e., the function should have
fail-omission failures only. This makes it easy for the system to timely
detect the absence of outputs and react accordingly. Through years,
safety monitors or safety wrappers have been applied to different
functions with beneficial effects on the non-functional (either safety
or security {[}13{]}, {[}17{]}) behavior of the component and the
encompassing systems. As a result, methodologies, techniques, and
industrial applications of safety monitors were largely applied to
different functional components and became solid literature with poor
research-wise interest.

However, the last twenty years saw a growing interest in developing
functional components that (partially) rely on Machine Learning (ML)
algorithms that perform classification (\emph{classifiers} in the
paper). Classifiers can model one or more expected behaviors of a system
or component and detect deviations that may be due to the occurrence of
faults or attacks, and perform error detection, intrusion detection,
failure prediction, or out-of-distribution detection {[}4{]}, {[}19{]},
{[}34{]}, to name a few. Straightforwardly, academia, industry, and also
National governments hugely invested in methodologies, mechanisms, and
tools to embed classifiers into ICT systems, including safety-critical
ones. However,
classifiers may predict a wrong class for a given data point, which is
typically called a \emph{misclassification.} This is a well-known
limitation to the adoption of classifiers to operate safety-critical
functions, requiring countermeasures that avoid or mitigate the
potential cascading effects of misclassifications to the encompassing
system. A fail-omission classifier would either produce trusted outputs
or omit them {[}30{]}. Clearly, this approach is different from building
a classifier that never outputs misclassifications, which is unrealistic
to assess at the state of the art due to the dimension of the input
space and the unpredictable behavior with inputs close to the decision
boundaries {[}34{]}.

This paper uses uncertainty measures that quantify the confidence in the
classification to craft a safety wrapper for black-box
classifiers. Uncertainty measures {[}2{]}, {[}10{]}, {[}12{]}, {[}35{]}
analyze inputs and/or predictions of the classifier and provide a
quantitative confidence evaluation. Their goal 
is to quantify the uncertainty of the classifier's predictions such that
there is significantly different uncertainty between i) the predictions
that turn out to be correct, and ii) those that turn out to be
misclassifications. In case of high uncertainty, the wrapper should omit
the output. We convey the observations above to design, implement and
evaluate a Safety wraPper thROugh ensembles of UncertainTy measures
(SPROUT). SPROUT wraps a classifier, computes different uncertainty
measures, and produces a binary confidence score to suspect
misclassifications and decide whether the prediction can be safely
propagated to the encompassing system, or if it should be omitted. It
can be widely applied because the wrapped classifier is seen as a
black-box: internal details do not need to be disclosed. 

More in detail, this paper summarizes techniques to compute the uncertainty in the predictions of classifiers, and considers a total of 9 uncertainty measures that can be instantiated with different parameters' values depending on the needs of the user. These allow to introduce the novel contributions of the paper, which we summarize in three items:

\begin{itemize}
\item
  discuss the application of safety wrappers (or safety monitors) to complement classifiers, converting misclassifications into omissions, and the implications it has in the overall classification process and for the encompassing system;
\item
  describe our Safety wraPper thROugh ensembles of UncertainTy measures
  (SPROUT) for black-box classifiers, which builds upon the discussion above. SPROUT is easy to use, adapts to any classifier, is publicly available at {[}5{]} and available as PIP package;
\item
  show how SPROUT is capable of detecting a huge fraction of the misclassifications of supervised classifiers, even omitting \emph{all} misclassifications in specific cases.
\end{itemize}

The paper is organized as follows. Section 2 reviews safety wrappers and the impact
they have on failure modes of critical systems. Section 3 describes
uncertainty measures, allowing Section 4 to
design SPROUT. Section 5 shows how to implement and exercise SPROUT.
Section 6 discusses the preliminary assessment, letting Section 7 list threats to validity, and Section 8 concludes the paper.

\section{Safety Wrappers for Machine Learning}

\subsection{On Misclassifications of Machine Learners}

Decades of research and practice on ML provided us with plenty of
classifiers that are meant to always output a prediction. Supervised
classifiers {[}4{]}, {[}15{]}, {[}20{]}, and particularly Deep Learners
{[}8{]}, {[}9{]} were proven to achieve excellent classification
performance in many domains: they learn their model using data points
collected i) during normal operation of the system, and ii) when errors,
attacks or failures activate; those data points are then labelled
accordingly.

More formally, a classifier \emph{clf} first devises a mathematical
model from a training dataset {[}4{]}, which contains a given amount of
data points. Each data point \emph{dp} contains a set of \emph{f}
feature values, where each feature value is an image pixel / channel or
a floating point number \emph{dp}\textsubscript{j} with $0 \leq j < f$ and describes a specific input of the
classification problem. Once the model is learned, it can be used to
predict the label \emph{dp\_label} of a new data point, different from
those in the training dataset. The classification performance is usually
computed by applying \emph{clf} to data points in a test dataset and
computing metrics such as \emph{accuracy} {[}31{]}, i.e., the percentage of correct
predictions of a classifier \emph{clf} over all predictions. Noticeably, \emph{1 - accuracy}
quantifies the misclassification probability by difference.

\subsection{Failure Modes of Classifiers and Safety Wrappers}

\begin{table}[b]
\begin{center}
{\caption{Probabilities $\alpha_w$, $\epsilon_w$, $\phi_c$, $\phi_m$ for outputs of SW(clf) and compound probabilities.}\label{Tab2}}
\setlength\tabcolsep{3pt} 
\begin{tabular*}{\linewidth}{c|cc|c}
\hline
\hline
clf behavior $\rightarrow$ & \multirow{2}{*}{Correct Classification} & \multirow{2}{*}{Mislassification} & \multirow{2}{*}{Sum} \\
SW(clf) behavior $\downarrow$ & & & \\
\hline
Not Omitted & $\alpha_w$ & $\epsilon_w$ & $1-\phi$ \\
Omitted & $\phi_c$ & $\phi_m$ & $\phi$ \\
\hline
Sum & $\alpha$ & $\epsilon$ & 1 \\
\hline
\hline
\end{tabular*}
\end{center}
\end{table}

Classifiers are typically meant to provide a best-effort prediction of
the class of input data according to the information they have, i.e.,
the input data and its features. As a result, classifiers sometimes
``bet'' on a prediction they are unsure of: in these cases, their
accuracy may drop significantly. It turns out evident that this
best-effort behavior does not pair well with safety-critical systems,
which require guarantees of component and system-level behaviors.

It would be beneficial to change the failure semantics of classifiers
from uncontrolled content failures (i.e., misclassifications) to
omission failures. Fail-controlled components {[}40{]} often rely on safety wrappers or
monitors {[}13{]}, {[}17{]}, {[}30{]}. Safety wrappers are intended to
complement an existing critical component or task by continuously
checking invariants, or processing additional data to detect dangerous
behaviors and block the erroneous output of the component before it is
propagated through the system. Safety wrappers for classifiers should
perform runtime monitoring and aim at detecting the misclassifications
of the classifier itself. Regardless of how it is implemented, a safety
wrapper \emph{SW(clf)} transforms a classifier \emph{clf} which has $0 \leq
\alpha \leq 1$ accuracy and a misclassification probability $0 \leq \epsilon = (1 - \alpha) \leq 1$, into a component that has:

\begin{itemize}
\item
  accuracy $\alpha_w \leq \alpha$;
\item
  omission probability $ 0 \leq \phi \leq 1$. The SW(clf) may omit
  misclassifications ($\phi_m$, desirable and to be
  maximized), or correct predictions ($\phi_c$, to be
  minimized). Overall, $\phi = \phi_m + \phi_c$, and
  $\alpha_w = \alpha - \phi_c $;
\item
  residual misclassification probability $\epsilon_w$, $0 \leq \epsilon_w \leq \epsilon \leq 1$; overall, $\epsilon_w = \epsilon - \phi_m$.
\end{itemize}

All those probabilities are sketched in Table 2. Ideally,
\emph{SW(clf)} has almost the same accuracy as \emph{clf} (i.e.,
$\alpha_w \approx \alpha$, or $\phi_c \approx 0$), a substantially
lower residual misclassification probability, $0 \approx \epsilon_w << \epsilon$, and an omission probability close to $\epsilon$
thus $\phi \approx \epsilon$. A \emph{SW(clf)} will never have better accuracy than \emph{clf}; however, it will transform most of the misclassifications,
which are hardly predictable, detectable and manageable, into omissions.

\subsection{Related Work and Motivation}

Recently, there have been few studies that specifically aim at building
safety wrappers for classifiers {[}3{]}, {[}33{]}, {[}36{]}, {[}37{]}.
The paper {[}3{]} ran a k-nearest neighbor classifier in parallel to a
Deep Neural Network (DNN) to detect misclassifications. The paper
{[}33{]} conducted an active monitoring of the behavior and the
operational context of the data-driven system based on distance measures
of the Empirical Cumulative Distribution Function, and used them as
triggers for the safety wrapper. The work {[}36{]} used probabilistic
neural networks to model predictive distributions and thus estimate
misclassifications thanks to adversarial training. This technique
performed well for image classifiers. Lastly, {[}37{]} proposed a
lightweight monitoring architecture to enhance the model robustness
against different unsafe inputs, especially those due to adversarial
attacks to neural networks. The logic to detect misclassifications
revolved around an analysis of activation patterns of neurons in the
layers of a specific neural network, which authors showed to be
distinguishable in case of an adversarial input. It is worth mentioning that existing safety wrappers above are
classifier-specific (i.e., {[}3{]}, {[}10{]}, {[}37{]}), often rely on
extensive knowledge of the classifier (e.g., {[}37{]} requires the
structure of the DNN to be disclosed), require the implementation of
complex and multi-step processes {[}33{]}, or apply only to specific
types of input data (e.g., {[}12{]}, {[}36{]} are specifically crafted
for image classifiers). Instead, we seek for an approach which i) applies to any classifier, which is seen as a black-box; ii) is easy to automatize, adopt and exercise to novel datasets or systems; iii) applies to binary and multi-class classification problems, and iv) does not have any constraint on the input data and as such works with tabular and image datasets.

\section{Uncertainty Measures for Classifiers}

This section summarizes uncertainty measures that were
previously applied to compute the confidence in the prediction of
classifiers.

\begin{table*}
\begin{center}
{\caption{Summary of Uncertainty Measures used in this study}\label{Tab1}}
\setlength\tabcolsep{2.3pt} 
\begin{tabular*}{\linewidth}{ccccccc}
\hline
\hline
UM \# & \specialcell[]{Name of the \\ Uncertainty Measure} & \specialcell[]{Needs \\ Offline Setup} & \specialcell[]{Uses \\ Input Data} & \specialcell[]{Uses Classifier \\ Output} & \specialcell[]{Uses \\ Classifier} & Parameters of the Measure \\
\hline
UM1 & Confidence Intervals & \checkmark & \checkmark &  &  & w: confidence level \\
UM2 & Maximum Likelihood &  &  & \checkmark &  & - \\
UM3 & Entropy of Probabilities &  &  & \checkmark &  & - \\
UM4 & Bayesian Uncertainty & \checkmark & \checkmark &  &  & - \\
UM5 & Combined Uncertainty & \checkmark & \checkmark & \checkmark &  & chk\_c: classifier to check agreement with \\
UM6 & MultiCombined Uncertainty & \checkmark & \checkmark & \checkmark &  & CC: classifiers to check agreement with \\
UM7 & Feature Bagging & \checkmark & \checkmark &  &  & bagC: classifier to build bagger set \\
UM8 & Neighbourhood Agreement &  & \checkmark &  & \checkmark & k: number of relevant neighbors \\
UM9 & Reconstruction Loss & \checkmark & \checkmark &  &  & layers: structure of the AutoEncoder \\
\hline
\hline
\end{tabular*}
\end{center}
\end{table*}

\subsection{Related Works on Uncertainty Measures and their Limitations}

Research usually aims at minimizing the probability of misclassifications, thus maximizing
accuracy. However, trusting each individual prediction of a classifier,
to the extent that the prediction can be propagated towards the
encompassing system and used in a (safety-)critical task, is a different
problem that is still open {[}12{]}. Researchers and practitioners are
actively investigating ways to understand if classifiers' predictions
are correct, or if they are misclassifications. The most relevant
research results on uncertainty measures are very recent (the last 5
years), which demonstrates the recent emergence and timeliness of the
topic. Uncertainty is often {[}38{]} referred to as a combination of \emph{aleatoric} and
\emph{epistemic} uncertainty. The former refers to the notion of
randomness, that is, the variability in the outcome of an experiment
which is due to inherently random effects e.g., coin-flip. The latter
describes uncertainty due to a lack of knowledge of any
underlying random phenomenon. In other words, \emph{epistemic
uncertainty refers to the reducible part of the (total) uncertainty,
whereas aleatoric uncertainty refers to the irreducible part} {[}38{]}.
Uncertainty measures quantify the epistemic uncertainty, and can hardly
provide useful information to estimate aleatoric uncertainty.

Uncertainty can be statistically estimated through confidence intervals {[}1{]} or using the Bayes theorem {[}2{]}. Works
as {[}11{]} estimate uncertainty by using ensembles of neural networks:
scores from the ensembles are combined in a unified measure that
describes the agreement of predictions and quantifies uncertainty. In {[}10{]}, {[}18{]}, authors
processed \emph{softmax} probabilities of neural networks to identify
misclassified data points. A new proposal came from {[}12{]} and {[}3{]}, where
authors paired a k-Nearest Neighbor classifier with a neural network to
compute uncertainty. The work
{[}34{]} computed the cross-entropy on the \emph{softmax} probabilities
of a neural network, and used it to detect
out-of-distribution input data that likely misclassified.

Uncertainty measures above compute either classifier-specific or
classifier-independent quantities. However, classifier-specific
uncertainty may not always be a meaningful indicator of
misclassifications since ``\emph{neural networks which yield a piecewise
linear classifier function {[}\ldots{]} produce almost always high
confidence predictions far away from the training data}'' {[}32{]}.

\subsection{Quantitative Measures to Compute Uncertainty}

This work focuses on uncertainty measures that are not
classifier-specific, but instead have a generic formulation that pairs
well with any classifier, which is seen as a black-box. This allows
avoiding classifier-specific uncertainty, which may be misleading
{[}32{]}. Table 1 summarizes a total of 9 uncertainty measures UM1 to
UM9, which process at least one of: i) input data \emph{dp}, ii) class
prediction \emph{dp\_prob}. Importantly, all measures but UM2, UM3 and UM8 require training data for set-up, and all measures but UM2, UM3, UM4 are parametric, meaning that different values of parameters may be employed to craft different instances of the same measure.

\textbf{UM1: Confidence Interval} A confidence interval defines the statistical distribution underlying the value of a feature and thus provides a range, constrained to the parameter $0 \leq w \leq 1$, in which feature values are expected to fall. The confidence level $w$ represents the long-run proportion of feature values (at the given confidence level) that theoretically contain the true value of the feature {[}41{]}. UM1 measures how many feature values falls inside their confidence interval. The higher the UM1, the more feature values of \emph{dp} are outside their confidence interval, which indicates high uncertainty in the prediction.

\textbf{UM2: Maximum Likelihood} Given \emph{dp\_prob} produced by a
classifier for a given \emph{dp}, we identify UM2 as the maximum
probability of \emph{dp}\_\emph{prob}. The higher the UM2, the more
uncertain the output of the classifier {[}18{]}.

\textbf{UM3: Entropy of Probabilities} We retrieve the \emph{dp\_prob}
produced by a classifier for a given \emph{dp} and we compute UM3 using
\emph{db\_prob} entropy {[}10{]}. The higher the UM3, the more uncertain
the classifier: a \emph{dp\_prob} array with constant values (i.e., all
classes have the same probability) generates the highest UM3 of 1.

\textbf{UM4: Bayesian Uncertainty} This measure uses a Naïve Bayes
process to estimate the probability that the input data point \emph{dp}
belongs to each of the possible \emph{c} classes {[}2{]}. Briefly, this
process applies Bayes\textquotesingle{} theorem assuming strong (i.e.,
naive) independence between the features. As such, UM4 may not apply to
many classification problems, especially those dealing with images,
where a pixel (feature) clearly depends on its surrounding pixels.

\textbf{UM5: Combined Uncertainty} UM5 uses a classifier \emph{chk\_c}
that acts as a checker of the main classifier \emph{clf}. UM5 has
positive sign if \emph{clf} and \emph{chk\_c} agree on the predicted
class, negative otherwise. The absolute value of UM5 is quantified
according to the entropy (UM3) in the results of \emph{chk\_c}. UM5
ranges from -1 to 1. UM5 = 1 translates to high confidence that the
prediction of \emph{clf} is correct, UM5 = -1 means high confidence that
the prediction is a misclassification, letting UM5 = 0 show maximum
uncertainty.

\textbf{UM6: Multi-Combined Uncertainty} UM6 computes uncertainty
relying on more than one checker. UM6 uses a set CC of \emph{ncc}
checking classifiers, computes UM5 for each $chk_c \in CC$ with
respect to \emph{clf}, and averages the results. The more checking
classifiers in CC agree with \emph{clf}, the higher the UM6.

\textbf{UM7: Feature Bagging} UM7 exploits the concept of bagging
{[}16{]}, a method for generating multiple versions of a classifier
\emph{bagC}: each instance of \emph{bagC} is trained using different
subsets of the original training set, and decides using restricted
knowledge. Should classifiers predict different classes for a given data
point \emph{dp}, UM7 would have low value and predictions should be
treated with high uncertainty.

\textbf{UM8: Neighbor Agreement} UM8 finds the \emph{k} nearest
neighbors {[}15{]} of a data point dp. Then, it classifies \emph{dp} and
its \emph{k} neighbors using \emph{clf}: the more neighbors are assigned
to the same class predicted for dp, the higher the UM8. The lower the
value, the more disagreement in classifying neighboring data points to
\emph{dp}. This means that the input data point \emph{dp} lies in an
unstable region of the input space, which translates to high uncertainty
(low UM8) in the prediction.

\textbf{UM9 Reconstruction Loss} Reconstruction loss quantifies to what
extent the input data point is an unseen, out-of-distribution data point
{[}19{]}, and as such it is likely to generate misclassifications. We
compute UM9 through the reconstruction error of autoencoders, which are
unsupervised neural networks composed of different \emph{layers} to learn
efficient encodings of the input data. A low UM9 value instead indicates
that \emph{dp} belongs to an expected distribution and as such is likely
to be correctly classified.

\section{SPROUT: a Safety wraPper thROugh ensembles of UncertainTy measures}

This section describes our safety wrapper for black-box
classifiers, binary or multiclass, which works with tabular and image data.

\subsection{Safety Wrappers for Black-Box Classifiers}

Figure 1a depicts a typical classifier: the input data, and the features
contained therein, are fed into a classifier \emph{clf} that predicts a
class label \emph{dp\_label} for that specific input data \emph{dp}.
This classification process always outputs a class label that is then
provided to the encompassing system, has $\alpha$ accuracy and $\epsilon = 1 - \alpha$ 
misclassification probability. In this scenario, all the
misclassifications are content failures. Figure 1b still feeds the input
data to the classifier, which predicts the class \emph{dp\_label} for an
input data \emph{dp}. However, the adoption of a safety wrapper
\emph{SW(clf)} provides the input data and the class prediction of the
\emph{clf} to a \emph{misclassification detector}, which outputs a
binary confidence score {[}30{]} (BCS) to decide if the class prediction
is detected to be a misclassification. In this case, the wrapper omits
the output (with probability $\phi$); otherwise, the class prediction gets
forwarded to the encompassing system, is correct with probability
$\alpha_w$ and is a misclassification with probability
$\epsilon_w$. There is still a residual probability
$\epsilon_w$ of content failure, while $\phi_m = \epsilon - \epsilon_w$ misclassifications are instead going to be omitted
thanks to the safety wrapper. Noticeably, insights of \emph{clf} do not
need to be disclosed for detecting misclassifications: as a result,
\emph{clf} is treated as a black-box classifier.

\begin{figure}[b]
\centerline{\includegraphics[height=1.2in]{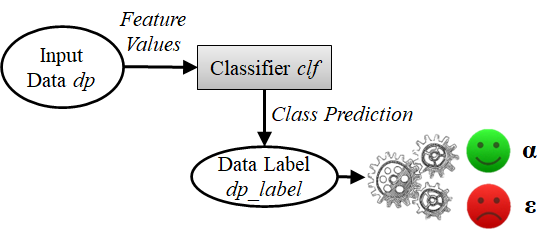}}
\caption*{a) Correctly predicts class with $\alpha$ accuracy and $\epsilon = 1 - \alpha$ misclassification probability}
\centerline{\includegraphics[height=1.1in]{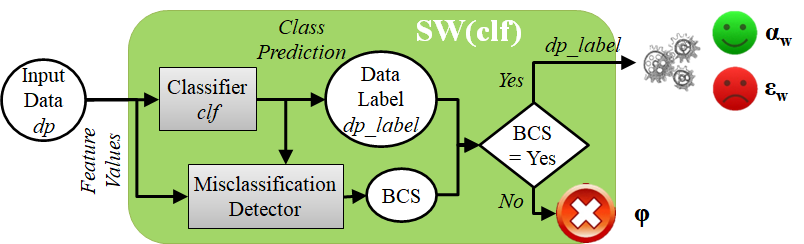}}
\caption*{b) Correctly predicts class with $\alpha_w$ accuracy, $\phi$ omission probability and $\epsilon_w$ residual misclassification probability.}
\caption{Classifier component (up, Figure 1a) and a classifier inside a safety wrapper (down, Figure 1b).}
\label{Fig1}
\end{figure}

The existence of a function to generate a \emph{dp\_label} and provide
the output probabilities of the classifier is the only assumption we
require for wrapping any classifier in such a \emph{SW(clf)} wrapper.
Note that commonly used frameworks for machine learning (to name a few:
\emph{scikit-learn, xgboost, pyod, tensorflow, pytorch}), expose such
interfaces; therefore \emph{SW(clf)} is virtually applicable to any
classifier without requiring compliance with restrictive assumptions.

Another important observation regards the applicability of SPROUT to any classifiers, regardless of the domain e.g., image classifiers or classifiers for tabular data, or the specific algorithm to be used, either a DNN, a tree-based classifier, a statistical classifiers, or any other binary or multi-class classifier. 

\begin{table*}
\begin{center}
{\caption{Example of uncertainty measures and misclassification flag using a supervised classifier on a specific dataset.}\label{Tab3}}
\setlength\tabcolsep{9pt} 
\begin{tabular*}{\linewidth}{cccccccccccc}
\hline
\hline
UM1 & UM2 & UM3 & UM4 & UM5 & UM6\_ST & UM6\_NB & UM6\_TR & UM7 & UM8 & UM9 & Misc flag \\
\hline
0.22 & 1.00 & -0.17 & 0.47 & 1.00 & 1.00 & 0.90 & 1.00 & 0.47 & 1.00 & 1.00 & correct \\
0.43 & 0.39 & -0.16 & 0.57 & -0.21 & 0.40 & 0.90 & 0.55 & 0.57 & -0.21 & 0.4 & misc \\
0.32 & 0.99 & -0.15 & -0.41 & 0.99 & 0.94 & 0.69 & 1.00 & -0.41 & 0.99 & 0.94 & correct \\ 
\hline
\hline
\end{tabular*}
\end{center}
\end{table*}

\subsection{A Misclassification Detector for SPROUT}

The misclassification detector for SPROUT is structured as follows.

SPROUT computes multiple uncertainty measures for each input data and /
or the corresponding classifier output: the choice of which uncertainty
measures should be computed is of utmost importance {[}32{]}. Some
uncertainty measures may make SPROUT detect most of the
misclassifications (thus the residual misclassification probability
$\epsilon_w$ would be very low) at a cost of many omissions, i.e.,
$\phi >> 0, \alpha_w << \alpha$. Conversely, other measures may build a SPROUT wrapper that has
optimal accuracy ($\alpha_w \approx \alpha$, or $\phi_c \approx 0$),
but rarely omits outputs ($\phi \approx 0$) and fails in detecting many
misclassifications ($\epsilon_w \approx \epsilon$, or $\phi_m \approx 0$)
making its behavior similar to the regular \emph{clf}. We tackle this
problem by relying on multiple uncertainty measures amongst those
presented in Section 2. Remember that several measures are parameter-dependent and as
such can be instantiated multiple times and have a different behavior;
this is the case of UM1, UM5, UM6, UM7, UM8 and UM9. The choice of parameters depends on the structure of the input data (e.g., tabular or image data), the type of the classification task (i.e., multi-class or binary) or other specific user needs.

Then, a binary adjudicator processes the ensemble of floating point values computed using each
uncertainty measure to output a unique
BCS. This binary adjudicator can be implemented with thresholds,
invariants, custom rules {[}14{]}, or as a binary classifier, providing many degrees of freedom in finding the ideal function to
combine ensembles of uncertainty measures into a unified BCS, even
implementing non-linear decision functions. The resulting misclassification detector will implement a stacking meta-learner, with uncertainty measures at the base level, and a binary adjudicator at the meta-level {[}60{]}. Obviously, the classifier
that implements each binary adjudicator is decoupled from the classifier
\emph{clf} used for classification.

\section{Exercising SPROUT}

This section details the experimental campaign to test SPROUT in detecting misclassifications of supervised classifiers.

\subsection{Experimental Methodology and its Inputs}

As a data baseline, we gather 33 public datasets: 11 datasets (i.e.,
NSL-KDD {[}44{]}, ISCX12 {[}43{]}, UNSW-NB15 {[}46{]}, UGR16 {[}50{]}, NGIDS-DS and ADFANet {[}48{]}, AndMal17 {[}49{]}, CIDDS-001 {[}45{]}, CICIDS17 and CICIDS18 {[}47{]}, SDN20 {[}51{]}) of network intrusion detection, datasets of sensor spoofing attacks to 10
different biometric traits summarized in {[}22{]} including Fingerprint {[}52{]}), Hand Gesture {[}56{]}), Electrodermal Activity {[}54{]}), Heart Rate {[}55{]}), Human Gait {[}57{]}), Keystroke {[}53{]}), Voice {[}58{]}), Face {[}59{]}), 7 BackBlaze and BAIDU datasets related to hardware monitoring for failure prediction {[}24{]}, {[}25{]}, 3 datasets related to IoT systems (ScaniaTrucks, MechFailure, Iot-IDS ) {[}27{]}, {[}26{]}, {[}28{]}, MNIST and Fashion-MNIST image datasets {[}6{]}, {[}7{]}.

We exercise the following 8 supervised classifiers that apply to both tabular and image data, and fit binary and multi-class classification: Decision Tree
(DT), Random Forests (RF), eXtreme Gradient Boosting (XGB), Logistic
Regression (LR), Naïve Bayes (NB), Linear Discriminant Analysis (LDA),
TabNet {[}8{]} and FastAI {[}9{]} neural networks. The neural networks
{[}8{]}, {[}9{]} are explicitly optimized for processing tabular data,
which are the majority of our datasets, but pair well also with image
datasets. Testing SPROUT with DNNs that are specifically tailored for image classification is something that we will discuss as future work. Regarding the choice of the hyper-parameters for those classifiers, we proceed as follows: we use the HyperOpt {[}42{]} library whenever possible (i.e., for all classifiers but FastAI and TabNet). We then let the hyperparameter optimizer that is embedded in FastAI to automatically tune its parameters. For TabNet we ran grid searches with 108
combinations of the following parameters and values: Learning rate
$\in [e^{-5}, e^{-3}, e^{-1}]$, Batch size $\in [512, 1024, 2048]$, Max Epochs
$\in [20, 50, 100]$, patience (for early stopping) $\in [5, 8]$, target metric $\in [mcc, accuracy]$.  

\begin{figure}[b]
\centerline{\includegraphics[width=\linewidth]{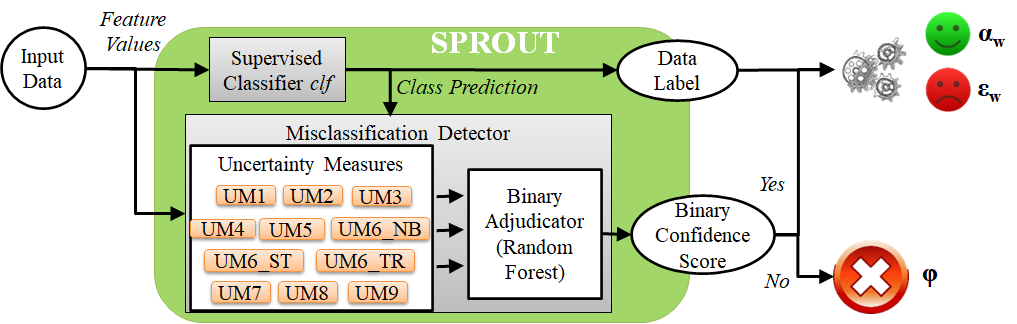}}
\caption{SPROUT wrapper defined and exercised in this paper.}
\label{Fig2}
\end{figure}

\begin{figure*}[h]
\centerline{\includegraphics[width=\textwidth]{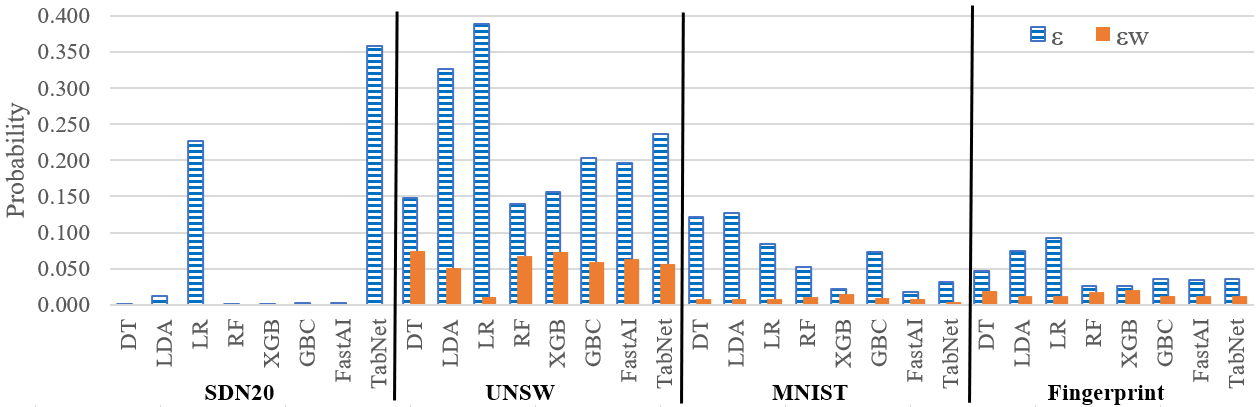}}
\caption{Misclassification probability $\epsilon$ of the classifier (striped bars), and residual misclassification probability $\epsilon_w$ (orange bars) of SPROUT for different 8 supervised classifiers exercised on SDN20, UNSW, MNIST and Fingerprint datasets.}
\label{Fig3}
\end{figure*}

We instantiate the following uncertainty measures:

\begin{itemize}
\item
  UM1 with w = 0.9.
\item
  UM2, UM3, and UM4, which do not have parameters.
\item
  UM5 with \emph{chk\_c} = XGB, which is a notoriously good classifier
  {[}29{]}.
\item
  UM6 with three different groups of checking classifiers. We indicate
  the three UM6 configuration as \emph{UM6\_ST)} \{NB, LDA, LR\},
  \emph{UM6\_TR}) \{DT, RF, XGB\}, and \emph{UM6\_NB)} \{GaussianNB,
  BernoulliNB, MultinomialNB, ComplementNB\}. UM6\_NB uses variants of
  the Naïve Bayes (NB) classifier.
\item
  UM7 with \emph{bagC} = DT, which has low computational complexity and
  has overall good classification performance. UM7 creates multiple
  instances of \emph{bagC}: therefore, a slow \emph{bagC} would make UM7
  take too much time.
\item
  UM8 with \emph{k} = 19: a prime k avoids ties in kNN searches
  {[}15{]}.
\item
  UM9 using 5 layers of the following size: ${f, f/2, f/4, f/2, f}$, being f the number of features in a dataset, which ranges from 4 (ADFANet dataset) to a maximum of 156 (ScaniaTrucks dataset).
\end{itemize}

Exercising each of the 8 classifiers on each of the 33 datasets and
computing uncertainty measures provides a total of 264 csv files that
are structured as shown in Table 3.

The reader would notice that we still did not discuss the implementation
of the binary adjudicator, which is a classifier and as such needs to be
trained itself. Therefore, we split the 264 csv files above into two
groups: uncertainty measures (plus the \emph{misc flag} label) computed
for the 8 classifiers on 29 datasets will build the training set of the
binary adjudicator, for a total of more than 13 million of labelled data
points. The remaining 1.8 million of data points, each containing the
uncertainty measures and \emph{misc\_flag} computed for \emph{SDN20,
UNSW, MNIST}, and \emph{Fingerprint} datasets, will be used as test set
for the binary adjudicator and to quantify performance of SPROUT in
detecting misclassifications.

We independently exercise Random Forest and XGB classifiers as binary
adjudicators, which are known to have excellent classification
performance for tabular data {[}29{]}: since Random Forests showed
better detection accuracy of misclassifications than XGB, we implement
the binary adjudicator of SPROUT as a Random Forest composed of 30
trees. This completes the definition and instantiation of SPROUT in
Figure 2.

Experiments have been executed on a Dell Precision 5820 Tower with an
Intel Xeon Gold 6250, GPU NVIDIA Quadro RTX6000 with 24GB VRAM, 192GB
RAM and Ubuntu 18.04, NVIDIA driver 450.119.03 with CUDA 11.0, and
required approximately three weeks of 24H execution with GPU support.

\subsection{A Library for Exercising SPROUT}

SPROUT is available at {[}5{]} and as PIP Python package. The package implements all uncertainty measures discussed in this paper and makes SPROUT ready for deployment in any case study. Many already trained models for binary adjudication are already available in the library, and are accompanied by details about the uncertainty calculators they need, statistics on its binary classification performance and on the importance each uncertainty calculator had in learning that model. Those information are not needed to run SPROUT but provide interesting details for explaining why SPROUT works as intended. 

Applying SPROUT to a brand new case study is very easy. Below we report a code snippet that shows a simple usage of SPROUT to wrap a supervised classifier from \emph{scikit-learn} using a pre-defined ($ecai_sup$) binary adjudicator which uses the architecture in Figure 2. We assume to have a labeled dataset that we split in two parts. The first part will be provided as input to the $load\_wrapper$ method that prepares a SPROUT wrapper according to the chosen model. Data is used to train the uncertainty measures, while the binary adjudicator is simply loaded from the repository. Then, we initialize and train a RF classifier, which we provide as input, alongside with unlabeled test data, to the $predict\_misclassifications$ function, which outputs i) a pandas DataFrame containing the values of uncertainty measures computed for all the test data points and the associated binary confidence score, and ii) the model of the classifier used for binary adjudication. Binary confidence scores are extracted as numpy array and used to compute $\phi$. If the test set is labeled, we can take advantage of labels to compute $\alpha_w$ and $\epsilon_w$ as shown in the last rows of Listing 1. Obviously, $y\_te$ labels will not available when deploying SPROUT in a real scenario.

\begin{python}
import sklearn as sk, numpy
from sprout.SPROUTObject import SPROUTObject
# We suppose having a dataset loaded as follows
#      x: a numpy matrix containing feature values
#      y: a numpy array containing dataset labels
#      label_tags: unique labels in y
x_tr, x_te, y_tr, y_te =
    sk.model_selection.train_test_split(
        x, y, test_size=0.5)
# Initializes an empty SPROUT wrapper.
so = SPROUTObject()
# Loads a specific model for binary adjudication
so.load_wrapper(model_tag='ecai_sup', x_train=x_tr,
     y_train=y_tr, la-bel_names=label_tags)
# Crafting classifier (can be any)
classifier = sk.ensemble.RandomForestClassifier()
classifier.fit(x_tr, y_tr)
# Suspects misclassifications of clf on test set
sp_df, bin_adj =
    so.predict_misclassifications(data_set=x_te,
           classifier=classifier)
# A numpy array of binary confidence scores
sprout_pred = sprout_df['pred'].to_numpy()
phi = 100*numpy.count_nonzero(sprout_pred == 0) 
           / len(sprout_pred)
# Computes alpha_w and eps_w 
#           (only if y_test is available)
aw = sum((1-sprout_pred)*(1-y_test)) /
          numpy.count_nonzero(sprout_pred == 0)
ew = 1 - phi - aw
\end{python}

\section{Results and Discussion}

\begin{figure}[b]
\centerline{\includegraphics[width=\linewidth]{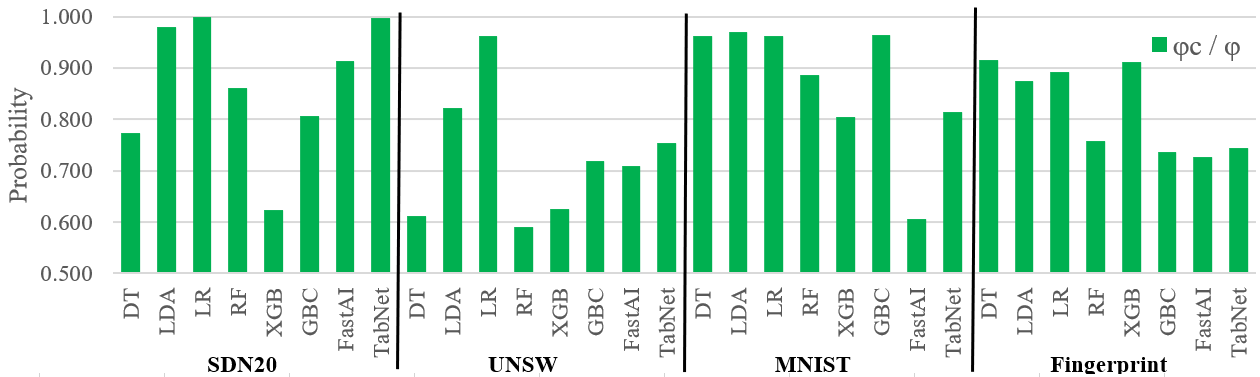}}
\caption{Rate $\phi_c  / \phi$ of omissions of misclassifications over omissions of SPROUT for different 8 supervised classifiers exercised on SDN20, UNSW, MNIST and Fingerprint datasets.}
\label{Fig4}
\end{figure}

\subsection{Detecting Misclassifications with SPROUT Wrappers}

Figure 3 reports a chart that compares the execution of each supervised
classifier with respect to its execution inside the SPROUT wrapper: blue
striped bars show the misclassification probability $\epsilon$ of \emph{clf},
while orange solid bars plot the residual misclassification probability
$\epsilon_w$ of \emph{SPROUT}. It turns out evident
that $\epsilon_w$ is always far lower than $\epsilon$ (i.e., orange bars hover
on the bottom of the plot and are always lower than 0.1, whereas the blue
bars may even reach 0.4), being extremely close to the optimum
$\epsilon_w \approx 0$ on SDN20 and MNIST datasets. There are cases in
which wrapping a \emph{clf} that has a high misclassification
probability $\epsilon$ may yield to the total absence of residual
misclassifications $\epsilon_w = 0$, which is an excellent result.
For instance, LR on SDN20 has $\epsilon = 0.2263$, meaning that more than 1 out of 5 predictions of \emph{clf} are misclassifications. Applying the
SPROUT wrapper leads to the total absence of residual misclassifications
($\epsilon_w = 0$), which is an excellent result safety-wise: the
output of SPROUT is either a correct misclassification or an omission.
As a drawback, SPROUT omits more than 1 out of 5 predictions of the
classifier ($\phi = 0.2264$), which is not desirable. This high omission
probability is a direct consequence of the high $\epsilon$ of LR classifier on
SDN20. When the \emph{clf} to be wrapped has high $\epsilon$ (as it happens in
the UNSW dataset), an high omission probability $\phi$ is unavoidable, even
when omitting all ($\epsilon_w = 0$) and only ($\phi_c = 0$) misclassifications.

On the downside, SPROUT may omit some outputs that were in fact going to
be correct classifications. We elaborate more on this aspect with the
aid of Figure 4, which plots the ratio of omissions of
misclassifications over all omissions i.e., $\phi_m / \phi$. The
higher the bars in the figure, the higher the $\phi_m / \phi$
ratio and the fewer omissions of correct classifications (i.e.,
$\phi_c$ is low). We elaborate more on the case of DT on the
UNSW dataset in Figure 4. In this case, only 60\% of omissions
correspond to misclassifications, which is not desirable: nevertheless, applying SPROUT has still a beneficial
impact on residual misclassifications ($\epsilon_w = \epsilon / 2$), but makes the accuracy $\alpha_w$ lower than the baseline ($\alpha = 0.852 > \alpha_w = 0.802$). A similar trend can be observed for RF on UNSW and for FastAI on MNIST.

\begin{table}[b]
\begin{center}
{\caption{Importance of uncertainty measures for binary adjudication (ranked by decreasing importance) and qualitative estimation on time required to compute each measure (H – high, M – medium, L – low, N – no impact, negligible time).}\label{Tab4}}
\setlength\tabcolsep{1pt} 
\begin{tabular*}{\linewidth}{cccccccccccc}
\hline
\hline
\specialcell[c]{Uncertainty \\ Measure} & \rotatebox[origin=c]{90}{UM7} & \rotatebox[origin=c]{90}{UM5} & \rotatebox[origin=c]{90}{UM6\_NB} & \rotatebox[origin=c]{90}{UM6\_ST} & \rotatebox[origin=c]{90}{UM6\_TR} & \rotatebox[origin=c]{90}{UM3} & \rotatebox[origin=c]{90}{UM1} & \rotatebox[origin=c]{90}{UM2} & \rotatebox[origin=c]{90}{UM9} & \rotatebox[origin=c]{90}{UM4} & \rotatebox[origin=c]{90}{UM8} \\
\hline
Importance & .289 & .189 & .138 & .128 & .128 & .036 & .032 & .027 & .017 & .010 & .004 \\
Time required & M & M & M & M & H & N & L & N & L & L & H \\
\hline
\hline
\end{tabular*}
\end{center}
\end{table}

\subsection{Importance of Uncertainty Measures}

Ultimately, we explore the impact each uncertainty measure has in
learning the model for binary adjudication and to detect
misclassifications with the aid of Table 4. The first row of the table
reports the importance each uncertainty measure (on the columns) has in
building the misclassification detector of SPROUT. Those scores are
computed through the \emph{feature\_importances\_} of
\emph{sklearn} \emph{Python} package, and sum up to 1. It turns out
evident that some uncertainty measure has marginal contribution for
binary adjudication and for detecting misclassifications. Particularly,
UM8 has the lowest feature importance and is almost entirely not
relevant for detecting misclassifications. Conversely, measures as UM7
and UM5 have the highest importance in building the models for binary
adjudication as they carry more information for detecting
misclassifications.

We also comment on the time needed to compute each uncertainty measure.
Table 4 reports a qualitative estimation for the time needed to
compute all uncertainty measures used by SPROUT. Some measures as UM2
and UM3 can be computed in negligible time and do not add any overhead
to the classification task. Measures as UM8 and UM6\_TR require heavy
computations which may significantly slowdown the execution of the
classification task. We are aware that the overhead generated by the
application of SPROUT may constitute an obstacle in systems which are
resource-limited or that have tight real-time deadlines. However, this
study primarily aims at building safety wrappers that can detect
misclassifications rather than optimizing speed. The reduction of the
timing overhead without affecting the other characteristics of SPROUT is
discussed as a future work.

\section{Threats to Validity and Reproducibility}

\textbf{Internal validity} is concerned with factors that may have influenced the results, but they have not been thoroughly considered in the study. First, public datasets are often collected from heterogeneous systems, may have been documented poorly, and are not under our control, but are of utmost importance for enabling reproducibility of this study. Second, classifiers have hyperparameters whose tuning critically affects results: therefore, we exercised sensitivity analyses for the main parameters of all classifiers considered in this study. Third, each classifier may encounter a wide variety of problems when learning a model for each dataset during training (e.g., under/overfitting, poor quality of features, feature selection to leave out noisy features). These events are mostly situational but may have a noticeable impact on the classification performance of a classifier. However, the reader should consider that this paper presents a safety wrapper that detects misclassifications of a black-box classifier, and therefore is not directly impacted by these problems that happened when training the main classifier.

\textbf{External validity}: we cannot claim the validity of this study for classifiers other than those that we used in this study i.e., DNNs for image classification, or unsupervised classifiers. In fact, this analysis is something the we will discuss shortly after as future works. Regarding the application domain of SPROUT, it fits any classification problem, but cannot be generalized easily to regression problems.

The usage of public data and public tools to run algorithms was a prerequisite of our analysis to allow \textbf{reproducibility} and to rely on proven-in-use data. We publicly shared scripts, methodologies and all metric scores, allowing any researcher or practitioner to repeat the experiments. We do not use any custom or private dataset: all dataset are referenced in the papers, and all code is available at {[}5{]}.

\section{Conclusions and Future Works}

This paper presented a safety wrapper to detect misclassifications of a
black-box classifier. SPROUT, our Safety wraPper thROugh ensembles of
UncertainTy measures, creates a wrapper around a classifier, either
binary or multi-class, and processes tabular or image input data. SPROUT
computes multiple uncertainty measures, providing quantitative data to
detect misclassifications of a classifier. Whenever a misclassification
is detected, SPROUT blocks the propagation of the output of the
classifier to the encompassing system: this way, a content failure of
the classifier is transformed into an omission failure, which can be
easily handled by the encompassing system. SPROUT wrappers
for supervised classifiers are available in the library available at {[}5{]}. Results in this paper show that SPROUT correctly detects the large majority of misclassifications of all the classifiers we used, and can even detect
all misclassifications of some classifiers (e.g., Logistic Regression on
the SDN20 dataset).

We are aware that we may have left out other uncertainty measures and
other groups of classifiers (i.e., unsupervised, neural networks for
image classification) from this study. Whereas the design and purpose of
SPROUT will not be affected by those additional measures and
classifiers, they may contribute to a more solid experimental analysis
which we plan as future work. In particular, we will craft SPROUT
wrappers for unsupervised classifiers and conduct additional experiments
that emphasize more on image classification, applying SPROUT to
pre-trained deep neural network models from the ImageNet model zoos
{[}39{]} of \emph{pytorch} and \emph{tensorflow}, and processing
well-known datasets other than those already considered in this study,
e.g., CIFAR-10 and ImageNet. As an additional but not least important
future work, we will focus on lowering the timing overhead introduced by
SPROUT wrappers with respect to a traditional classification process.
Uncertainty measures that individually introduce major overhead will be
evaluated to understand if i) they could be dropped without affecting
the behavior of wrappers, ii) their implementation could be optimized,
or iii) they could be replaced with faster alternatives.

\newpage

\ack This work was partially supported by project SERICS (PE00000014) under the MUR National Recovery and Resilience Plan funded by the European Union - NextGenerationEU and by the NextGenerationEU program, Italian DM737 – CUP B15F21005410003." 

\section*{References}

\begin{enumerate}
\def\labelenumi{[\arabic{enumi}]}
\item
  Meeker, W. Q., Hahn, G. J., \& Escobar, L. A. (2017). Statistical
  intervals: a guide for practitioners and researchers (Vol. 541). John
  Wiley \& Sons.
\item
  Krzanowski, W. J., et. Al. (2006). Confidence in classification: a
  bayesian approach. Journal of Classification, 23(2), 199-220.
\item
  Bilgin, Z., \& Gunestas, M. (2021). Explaining Inaccurate Predictions
  of Models through k-Nearest Neighbors. In ICAART (2) (pp. 228-236).
\item
  Bishop, C. M. (2006). Pattern recognition. Machine learning, 128(9).
\item
  SPROUT Repository on GitHub (online), 
  https://github.com/tommyippoz/SPROUT
\item
  LeCun, Y. (1998). The MNIST database of handwritten digits. (online)
  \url{http://yann.lecun.com/exdb/mnist/}.
\item
  Fashion-MNIST: a Novel Image Dataset for Benchmarking Machine Learning
  Algorithms. Han Xiao, Kashif Rasul, Roland Vollgraf. arXiv:1708.07747
\item
  Arik, S. Ö., \& Pfister, T. (2021, May). Tabnet: Attentive
  interpretable tabular learning. In Proceedings of the AAAI Conference
  on Artificial Intelligence, Vol. 35 (8), pp. 6679-6687.
\item
  Howard, J. et. Al. (2020). Fastai: a layered API for deep learning.
  Information, 11(2), 108.
\item
  Dan Hendrycks and Kevin Gimpel. A baseline for detecting misclassified
  and out-of-distribution examples in neural networks. arXiv preprint
  arXiv:1610.02136, 2016.
\item
  Lakshminarayanan, et. Al. Safety and scalable predictive uncertainty
  estimation using deep ensembles. In Advances in Neural Information
  Processing Systems, pp 6405--6416, 2017.
\item
  Jiang, H., Kim, B., Guan, M., \& Gupta, M. (2018). To trust or not to
  trust a classifier. Advances in neural information processing systems,
  31.
\item
  Pham, C., Estrada, Z., Cao, P., Kalbarczyk, Z., \& Iyer, R. K. (2014,
  June). Reliability and security monitoring of virtual machines using
  hardware architectural invariants. In 2014 44\textsuperscript{th}
  IEEE/IFIP Int. Conference on Dependable Systems and Networks (pp.
  13-24). IEEE.
\item
  Di Giandomenico, F., \& Strigini, L. (1990, October). Adjudicators for
  diverse-redundant components. Proc. 9\textsuperscript{th} Symposium on
  Reliable Distributed Systems (pp. 114-123). IEEE.
\item
  Cheung, K. L., \& Fu, A. W. C. (1998). Enhanced nearest neighbor
  search on the R-tree. AUM SIGMOD Record, 27(3), 16-21.
\item
  Breiman, L. (1996). Bagging predictors. Machine learning, 24(2),
  123-140.
\item
  Tiwari, A., Dutertre, B., Jovanović, D., de Candia, T., Lincoln, P.
  D., Rushby, J., \ldots{} \& Seshia, S. (2014, April). Safety wrapper
  for security. In Proceedings of the 3\textsuperscript{rd}
  international conference on High confidence networked systems (pp.
  85-94).
\item
  Fonseca, J. R., et. al. (2005). Uncertainty identification by the
  maximum likelihood method. Journal of Sound and Vibration, 288(3),
  587-599.
\item
  Xiao, Z., Yan, Q., \& Amit, Y. (2020). Likelihood regret: An
  out-of-distribution detection score for variational auto-encoder.
  Advances in neural information processing systems, 33, 20685-20696.
\item
  Kramer, M. A. (1991). Nonlinear principal component analysis using
  autoassociative neural networks. AIChE journal, 37(2), 233-243.
\item
  Nour Moustafa, Jill Slay. 2015. ``UNSW-NB15: a comprehensive data set
  for network intrusion detection systems''. In Military Communications
  and Information Systems Conference (MilCIS), 2015. IEEE, 1--6.
\item
  Zoppi, T., Gharib, M., Atif, M., \& Bondavalli, A. (2021).
  Meta-Learning to Improve Unsupervised Intrusion Detection in
  Cyber-Physical Systems. ACM Transactions on Cyber-Physical Systems
  (TCPS), 5(4), 1-27.
\item
  Ring, M., Wunderlich, S., Scheuring, D., Landes, D., \& Hotho, A.
  (2019). A survey of network-based intrusion detection data sets.
  Computers \& Security.
\item
  BackBlaze: BackBlaze Hard Drive Data (online)
  \url{https://www}.backblaze.com/b2/hard-drive-test-data.html
\item
  BAIDU: Baidu Smart HDD Competition (online)
  \url{https://www}.kaggle.com/drtycoon/baidu-hdds-dataset-2017/version/1
\item
  MechFailure: Machine Failure Prediction Competition (online),
  \url{https://www}.kaggle.com/c/machine-failure-prediction
\item
  Gondek C., e. al. (2016) Prediction of Failures in the Air Pressure
  System of Scania Trucks Using a Random Forest and Feature Engineering.
  In Advances in Intelligent Data Analysis XV. IDA 2016. Lecture Notes
  in Computer Science, vol 9897. Springer, Cham
\item
  IoT-IDS: IoT Intrusion (online)
  \url{https://ieee}-dataport.org/open-access/iot-network-intrusion-dataset\#files
\item
  Shwartz-Ziv, R., \& Armon, A. (2022). Tabular data: Deep learning is
  not all you need. Information Fusion, 81, 84-90.
\item
  Guérin, J., Ferreira, R. S., Delmas, K., \& Guiochet, J. (2022,
  October). Unifying evaluation of machine learning safety monitors. In
  2022 IEEE 33rd International Symposium on Software Reliability
  Engineering (ISSRE) (pp. 414-422). IEEE.
\item
  Lever, J. (2016). Classification evaluation: It is important to
  understand both what a classification metric expresses and what it
  hides. Nature methods, 13(8), 603-605.
\item
  Hein, M., Andriushchenko, M., \& Bitterwolf, J. (2019). Why relu
  networks yield high-confidence predictions far away from the training
  data and how to mitigate the problem. In Proc. of the IEEE/CVF
  Conference on Computer Vision and Pattern Recognition (pp. 41-50).
\item
  Aslansefat, K., et. al. (2020, September). SafeML: safety monitoring
  of machine learning classifiers through statistical difference
  measures. In International Symposium on Model-Based Safety and
  Assessment (pp. 197-211). Springer, Cham.
\item
  Wang, M., Shao, Y., Lin, H., Hu, W., \& Liu, B. (2022). Cmg: A
  class-mixed generation approach to out-of-distribution detection.
  Proceedings of ECML/PKDD-2022.
\item
  Shafaei, S., Kugele, S., Osman, M. H., \& Knoll, A. (2018, September).
  Uncertainty in machine learning: A safety perspective on autonomous
  driving. In International Conference on Computer Safety, Reliability,
  and Security (pp. 458-464). Springer, Cham.
\item
  Lakshminarayanan, B., Pritzel, A., \& Blundell, C. (2017). Simple and
  scalable predictive uncertainty estimation using deep ensembles.
  Advances in neural information processing systems, 30.
\item
  Rossolini, G., Biondi, A., \& Buttazzo, G. (2022). Increasing the
  Confidence of Deep Neural Networks by Coverage Analysis. IEEE
  Transactions on Software Engineering.
\item
  Hüllermeier, E., \& Waegeman, W. (2021). Aleatoric and epistemic
  uncertainty in machine learning: An introduction to concepts and
  methods. Machine Learning, 110(3), 457-506.
\item
  Model Zoo - Discover open source deep learning code and pretrained
  models (online), \url{https://modelzoo.co/} accessed: 2023-01-20
\item
  Avizienis, A., Laprie, J. C., Randell, B., \& Landwehr, C. (2004).
  Basic concepts and taxonomy of dependable and secure computing. IEEE
  transactions on dependable and secure computing, 1(1), 11-33.
\item 
  Hazra, A. (2017). Using the confidence interval confidently. Journal of thoracic disease, 9(10), 4125.
\item 
  Bergstra, J., Komer, B., Eliasmith, C., Yamins, D., \& Cox, D. D. (2015). Hyperopt: a python library for model selection and hyperparameter optimization. Computational Science \& Discovery, 8(1), 014008.

\item 
Ali Shiravi, Hadi Shiravi, Mahbod Tavallaee, and Ali A Ghorbani. 2012. Toward developing a systematic approach to generate benchmark datasets for intrusion detection. Computers \& Security 31, 3 (2012), 357–374.
\item 
Mahbod Tavallaee, Ebrahim Bagheri, Wei Lu, and Ali A Ghorbani. 2009. A detailed analysis of the KDD CUP 99 data set. In Computational Intelligence for Security and Defense Applications, 2009. CISDA 2009. IEEE Symposium on. IEEE, 1–6.
\item 
Ring, M., et. Al. (2017, June). Flow-based benchmark data sets for intrusion detection. In Proceedings of the 16th European Conference on Cyber Warfare and Security. ACPI (pp. 361-369).
\item 
Nour Moustafa, Jill Slay. 2015. “UNSW-NB15: a comprehensive data set for network intrusion detection systems”. In Military Communications and Information Systems Conference (MilCIS), 2015. IEEE, 1–6.
\item 
Sharafaldin, I., Lashkari, A. H., \& Ghorbani, A. A. (2018, January). Toward Generating a New Intrusion Detection Dataset and Intrusion Traffic Characterization. In ICISSP (pp. 108-116).
\item 
Haider, W., Hu, J., Slay, J., Turnbull, B. P., \& Xie, Y. (2017). Generating realistic intrusion detection system dataset based on fuzzy qualitative modeling. Journal of Network and Computer Applications, 87, 185-192.
\item 
Lashkari, A. H., et. Al. (2018, October). Toward Developing a Systematic Approach to Generate Benchmark Android Malware Datasets and Classification. In International Carnahan Conference on Security Technology (ICCST) (pp. 1-7). IEEE.
\item 
Maciá-Fernández, G., Camacho, J., Magán-Carrión, R., García-Teodoro, P., \& Theron, R. (2018). UGR ’16: A new dataset for the evaluation of cyclostationarity-based network IDSs. Computers \& Security, 73, 411-424.
\item 
Elsayed, M. S., Le-Khac, N. A., \& Jurcut, A. D. (2020). InSDN: A Novel SDN Intrusion Dataset. IEEE Access, 8, 165263-165284.
\item 
BIT – Biometrics Ideal Test, CASIA-FingerprintV5, http://biometrics.idealtest.org/
\item 
Adams, Warwick R. “High-accuracy detection of early Parkinson’s Disease using multiple characteristics of finger movement while typing.” PloS one 12.11 (2017): e0188226.
\item 
Koldijk, S., Sappelli, M., Verberne, S., Neerincx, M. A., \& Kraaij, W. (2014, November). The swell knowledge work dataset for stress and user modeling research. In Proceedings of the 16th international conference on multimodal interaction (pp. 291-298).
\item 
Philip Schmidt, Attila Reiss, Robert Duerichen, Claus Marberger, Kristof Van Laerhoven, “Introducing WESAD, a multimodal dataset for Wearable Stress and Affect Detection”, ICMI 2018, Boulder, USA, 2018
\item 
A. Memo, L. Minto, P. Zanuttigh,  “Exploiting Silhouette Descriptors and Synthetic Data for Hand Gesture Recognition”, STAG: Smart Tools \& Apps for Graphics, 2015
\item 
Vajdi, A., Zaghian, M. R., Farahmand, S., Rastegar, E., Maroofi, K., Jia, S., ... \& Bayat, A. (2019). Human Gait Database for Normal Walk Collected by Smart Phone Accelerometer. arXiv preprint arXiv:1905.03109.
\item 
Kaggle – Voice Recognition, Jeganathan Kolappan. https://www.kaggle.com/jeganathan/voice-recognition (online), accessed: 2022-11-20
\item 
Kaggle – Face Images with Marked Landmark Points, Omri Goldstein. https://www.kaggle.com/drgilermo/face-images-with-marked-landmark-points (online), accessed: 2022-11-20
\item 
Wolpert, D. H. (1992). Stacked generalization. Neural networks, 5(2), 241-259.
  
\end{enumerate}

\end{document}